# PubTator 3.0: an AI-powered Literature Resource for Unlocking Biomedical Knowledge

Chih-Hsuan Wei[1,1], Alexis Allot[1,†], Po-Ting Lai[1], Robert Leaman[1], Shubo Tian[1], Ling Luo[1], Qiao Jin[1], Zhizheng Wang[1], Qingyu Chen[1] and Zhiyong Lu[1,*]

[1] National Center for Biotechnology Information (NCBI), National Library of Medicine (NLM), National Institutes of Health (NIH), MD, 20894, Bethesda, USA
* To whom correspondence should be addressed. Tel: +1 301 594 7089; Email: zhiyong.lu@nih.gov

Present Address: Alexis Allot, The Neuro (Montreal Neurological Institute-Hospital), McGill University, Montreal, Quebec H3A 2B4, Canada
Present Address: Ling Luo, School of Computer Science and Technology, Dalian University of Technology, 116024, Dalian, China
Present Address: Qingyu Chen, Biomedical Informatics and Data Science, Yale School of Medicine, CT, 06510, New Haven, USA

**ABSTRACT**

PubTator 3.0 (https://www.ncbi.nlm.nih.gov/research/pubtator3/) is a biomedical literature resource using state-of-the-art AI techniques to offer semantic and relation searches for key concepts like proteins, genetic variants, diseases, and chemicals. It currently provides over one billion entity and relation annotations across approximately 36 million PubMed abstracts and 6 million full-text articles from the PMC open access subset, updated weekly. PubTator 3.0's online interface and API utilize these precomputed entity relations and synonyms to provide advanced search capabilities and enable large-scale analyses, streamlining many complex information needs. We showcase the retrieval quality of PubTator 3.0 using a series of entity pair queries, demonstrating that PubTator 3.0 retrieves a greater number of articles than either PubMed or Google Scholar, with higher precision in the top 20 results. We further show that integrating ChatGPT (GPT-4) with PubTator APIs dramatically improves the factuality and verifiability of its responses. In summary, PubTator 3.0 offers a comprehensive set of features and tools that allow researchers to navigate the ever-expanding wealth of biomedical literature, expediting research and unlocking valuable insights for scientific discovery.

---

[1] Joint Authors

**INTRODUCTION**

The biomedical literature is a primary resource to address information needs across the biological and clinical sciences (1), however the requirements for literature search vary widely. Activities such as formulating a research hypothesis require an exploratory approach, whereas tasks like interpreting the clinical significance of genetic variants are more focused.

Traditional keyword-based search methods have long formed the foundation of biomedical literature search (2). While generally effective for basic search, these methods also have significant limitations, such as missing relevant articles due to differing terminology or including irrelevant articles because surface-level term matches cannot adequately represent the required association between query terms. These limitations cost time and risk information needs remaining unmet.

Natural language processing (NLP) methods provide substantial value for creating bioinformatics resources (3-5), and may improve literature search by enabling semantic and relation search. In semantic search, users indicate specific concepts of interest (entities) for which the system has precomputed matches regardless of the terminology used. Relation search increases precision by allowing users to specify the type of relationship desired between entities, such as whether a chemical enhances or reduces expression of a gene. In this regard, we present PubTator 3.0, a novel resource engineered to support semantic and relation search in the biomedical literature. Its search capabilities allow users to explore automated entity annotations for six key biomedical entities: genes, diseases, chemicals, genetic variants, species, and cell lines. PubTator 3.0 also identifies and makes searchable 12 common types of relations between entities, enhancing its utility for both targeted and exploratory searches. Focusing on relations and entity types of interest across the biomedical sciences allows PubTator 3.0 to retrieve information precisely while providing broad utility (see detailed comparisons with its predecessor in Supplementary Table 1).

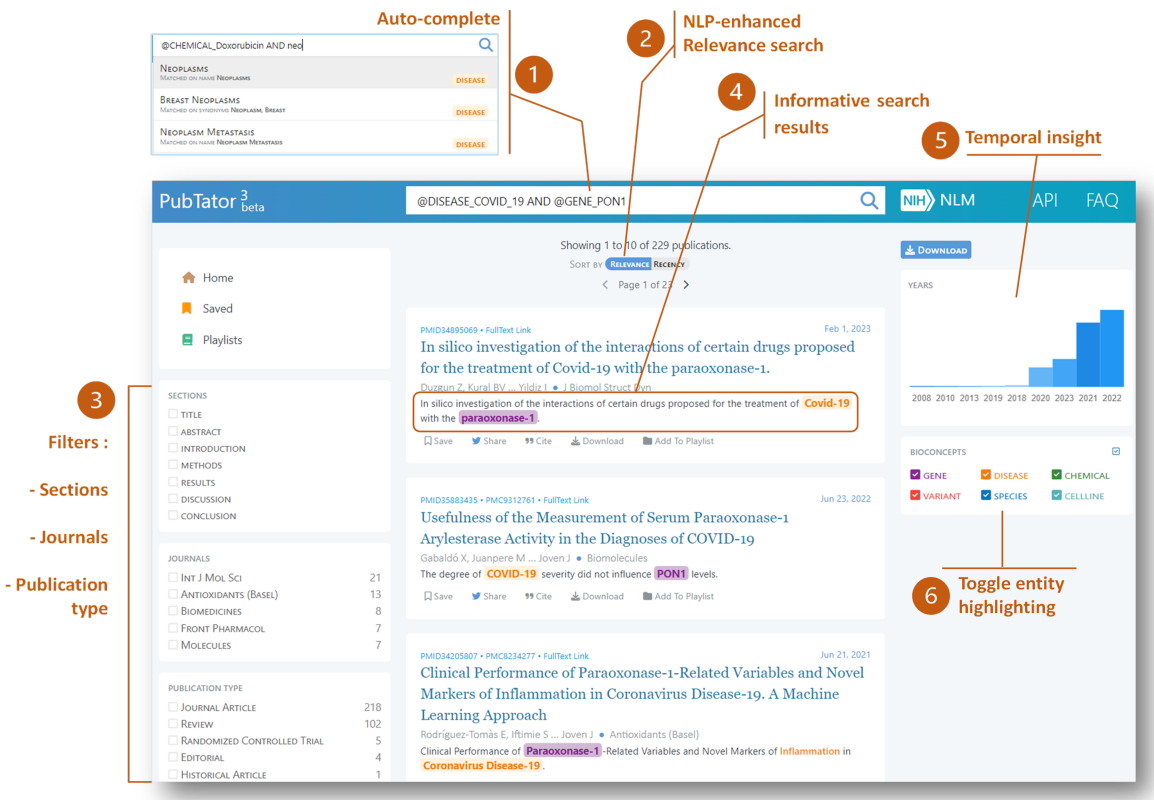

**Figure 1**. PubTator 3.0 system overview and search results page: 1. Query auto-complete enhances search accuracy and synonym matching. 2. Natural language processing (NLP)-enhanced relevance: Search results are prioritized according to the depth of the relationship between the entities queried. 3. Users can further refine results with facet filters—section, journal, and type. 4. Search results include highlighted entity snippets explaining relevance. 5. Histogram visualizes number of results by publication year. 6. Entity highlighting can be switched on or off according to user preference.

## SYSTEM OVERVIEW

The PubTator 3.0 online interface, illustrated in Fig. 1 and Supplementary Fig. 1, is designed for interactive literature exploration, supporting semantic, relation, keyword, and Boolean queries. An auto-complete function provides semantic search suggestions to assist users with query formulation. For example, it automatically suggests replacing either "COVID-19" or "SARS-CoV-2 infection" with the semantic term "@DISEASE_COVID_19". Relation queries – new to PubTator 3.0 – provide increased precision, allowing users to target articles which discuss specific relationships between entities.

PubTator 3.0 offers unified search results, simultaneously searching approximately 36 million PubMed abstracts and over 6 million full-text articles from the PMC Open Access Subset (PMC-OA), improving access to the substantial amount of relevant information present in the article full text (6). Search results are prioritized based on the depth of the relationship between the query terms: articles containing identifiable relations between semantic terms receive the highest priority, while articles where semantic or keyword terms co-occur nearby (e.g., within the same sentence) receive secondary priority. Search results are also prioritized based on the article section where the match appears (e.g., matches within the title receive higher priority). Users can further refine results by employing filters, narrowing articles returned to specific publication types, journals, or article sections.

PubTator 3.0 is supported by an NLP pipeline, depicted in Fig. 2A. This pipeline, run weekly, first identifies articles newly added to PubMed and PMC-OA. Articles are then processed through three major steps: 1. named entity recognition, provided by the recently developed deep-learning transformer model AIONER (7), 2. identifier mapping, and 3. relation extraction, performed by BioREx (8) of 12 common types of relations (described in Supplementary Table 2).

In total, PubTator 3.0 contains over 1.6 billion entity annotations (4.6 million unique identifiers) and 33 million relations (8.8 million unique pairs). It provides enhanced entity recognition and normalization performance over its previous version, PubTator 2 (9), also known as PubTator Central (Fig. 2B and Supplementary Table 3). We show the relation extraction performance of PubTator 3.0 in Fig. 2C and its comparison results to the previous state-of-the-art systems (10-12) on the BioCreative V Chemical-Disease Relation (13) corpus, finding that PubTator 3.0 provided substantially higher accuracy. Moreover, when evaluating a randomized sample of entity pair queries compared to PubMed and Google Scholar, PubTator 3.0 consistently returns a greater number of articles with higher precision in the top 20 results (Fig. 2D and Supplementary Table 4).

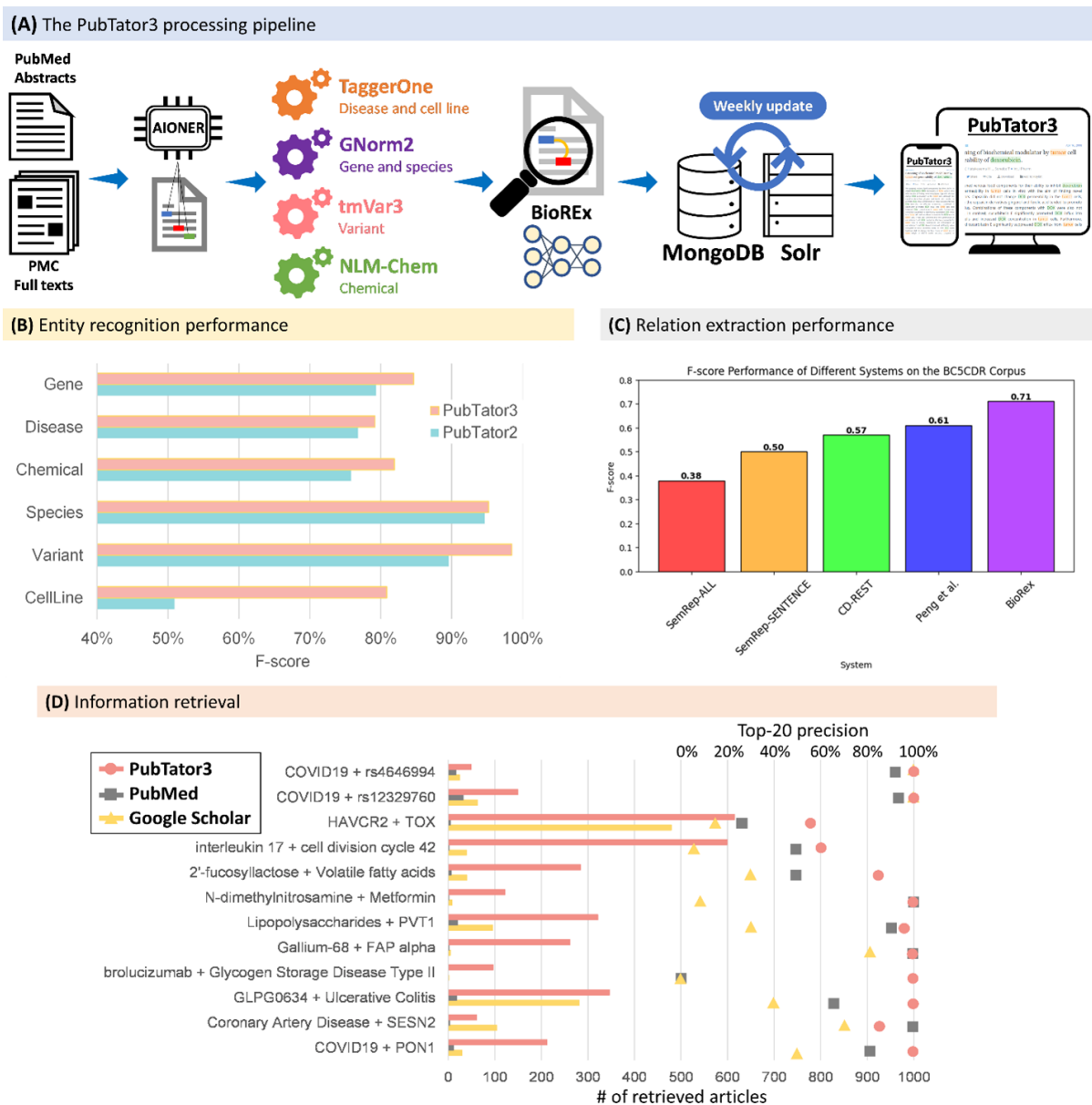

**Figure 2**. A. The PubTator 3.0 processing pipeline: AIONER (7) identifies six types of entities in PubMed abstracts and PMC-OA full-text articles. Entity annotations are associated with database identifiers by specialized mappers and BioREx (8) identifies relations between entities. Extracted data is stored in MongoDB and made searchable using Solr. B. Entity recognition performance for each entity type compared with PubTator2 (also known as PubTatorCentral) (13) on the BioRED corpus (14). C. Relation

extraction performance compared with SemRep (10) and notable previous best systems (11,12) on the BioCreative V Chemical-Disease Relation (13) corpus. D. Comparison of information retrieval for PubTator 3.0, PubMed, and Google Scholar for entity pair queries, with respect to total article count and top-20 article precision.

## MATERIAL AND METHODS

### Data Sources and Article Processing

PubTator 3.0 downloads new articles weekly from the BioC PubMed API (https://www.ncbi.nlm.nih.gov/research/bionlp/APIs/BioC-PubMed/) and the BioC PMC API (https://www.ncbi.nlm.nih.gov/research/bionlp/APIs/BioC-PMC/) in BioC-XML format (15). Local abbreviations are identified using Ab3P (16). Article text and extracted data are stored internally using MongoDB and indexed for search with Solr, ensuring robust and scalable accessibility unconstrained by external dependencies such as the NCBI eUtils API.

### Entity Recognition and Normalization / Linking

PubTator 3.0 uses AIONER (7), a recently developed named entity recognition (NER) model, to recognize entities of six types: genes/proteins, chemicals, diseases, species, genetic variants, and cell lines.  AIONER utilizes a flexible tagging scheme to integrate training data created separately into a single resource. These training datasets include NLM-Gene (17), NLM-Chem (18), NCBI-Disease (19), BC5CDR (13), tmVar3 (20), Species-800 (21), BioID (22), and BioRED (14). This consolidation creates a larger training set, improving the model's ability to generalize to unseen data. Furthermore, it enables recognizing multiple entity types simultaneously, enhancing efficiency and simplifying the challenge of distinguishing boundaries between entities that reference others, such as the disorder "Alpha-1 antitrypsin deficiency" and the protein "Alpha-1 antitrypsin." We previously evaluated the performance of AIONER on 14 benchmark datasets (7), including the test sets for the aforementioned training sets.  This evaluation demonstrated that AIONER's performance surpasses or matches previous state-of-the-art methods.

Entity mentions found by AIONER are normalized (linked) to a unique identifier in an appropriate entity database. Normalization is performed by a module designed for (or adapted to) each entity type, using the latest version. The recently-upgraded GNorm2 system (23) normalizes genes to NCBI Gene identifiers and species mentions to NCBI Taxonomy. tmVar3 (20), also recently upgraded, normalizes genetic variants; it uses dbSNP identifiers for variants listed in dbSNP and HGNV format otherwise. Chemicals are normalized by the NLM-Chem tagger (18) to MeSH identifiers (24). TaggerOne (25) normalizes diseases to MeSH and cell lines to Cellosaurus (26) using an improved normalization-only mode. These enhancements provide a significant overall improvement in entity normalization performance (Supplementary Table 2).

**Relation Extraction**

Relations for PubTator 3.0 are extracted by the unified relation extraction model BioREx (8), designed to simultaneously extract 12 types of relations across eight entity type pairs: chemical-chemical, chemical-disease, chemical-gene, chemical-variant, disease-gene, disease-variant, gene-gene, and variant-variant. Detailed definitions of these relation types and their corresponding entity pairs are presented in Supplementary Table 2. Deep-learning methods for relation extraction, such as BioREx, require ample training data. However, training data for relation extraction is fragmented into many datasets, often tailored to specific entity pairs. BioREx overcomes this limitation with a data-centric approach, reconciling discrepancies between disparate training datasets to construct a comprehensive, unified dataset.

We evaluated the relations extracted by BioREx using performance on manually annotated relation extraction datasets as well as a comparative analysis between BioREx and notable comparable systems. BioREx established a new performance benchmark on the BioRED corpus test set (14), elevating the performance from 74.4% (F-score) to 79.6%, and demonstrating higher performance than alternative models such as transfer learning (TL), multi-task learning (MTL), and state-of-the-art models trained on isolated datasets (8). For PubTator 3.0, we replaced its deep learning module, PubMedBERT (27), with LinkBERT (28), further increasing the performance to 82.0%. Furthermore, we conducted a comparative analysis between BioREx and SemRep (10), a widely used rule-based method for extracting diverse relations, the CD-REST (12) system, and the previous state-of-the-art system (11), using the BioCreative V Chemical Disease Relation corpus test set (13). Our evaluation demonstrated that PubTator 3.0 provided substantially higher F-score than previous methods.

**Programmatic Access and Data Formats**

PubTator 3.0 offers programmatic access through its API and bulk download. The API (https://www.ncbi.nlm.nih.gov/research/pubtator3/) supports keyword, entity and relation search, and also supports exporting annotations in XML and JSON-based BioC (15) formats and tab-delimited free text. The PubTator 3.0 FTP site (https://ftp.ncbi.nlm.nih.gov/pub/lu/PubTator3) provides bulk downloads of annotated articles and extraction summaries for entities and relations. Programmatic access supports more flexible query options; for example, the information need "what chemicals reduce expression of JAK1?" can be answered directly via API (e.g., https://www.ncbi.nlm.nih.gov/research/pubtator3-api/relations?e1=@GENE_JAK1&type=negative_correlate&e2=Chemical) or by filtering the bulk relations file. Additionally, the PubTator 3.0 API supports annotation of user-defined free text.

**Case Study I: Entity Relation Queries**

We analyzed the retrieval quality of PubTator 3.0 by preparing a series of 12 entity pairs to serve as case studies for comparison between PubTator 3.0, PubMed, and Google Scholar. To provide an equal comparison, we filtered Google Scholar results for articles not in PubMed. To ensure that the number of results would remain low enough to allow filtering Google Scholar results for articles not in PubMed, we identified entity pairs first discussed together in the literature in 2022 or later. We then randomly selected two entity pairs of each of the following types: Disease/Gene, Chemical/Disease, Chemical/Gene, Chemical/Chemical, Gene/Gene and Disease/Variant. The comparison was performed with respect to a snapshot of the search results returned by all search engines on May 19, 2023. We manually evaluated the top 20 results for each system and each query; articles were judged to be relevant if they mentioned both entities in the query and supported a relationship between them.

Our analysis is summarized in Fig. 2D, and Supplementary Table 4 presents a detailed comparison of the quality of retrieved results between PubTator 3.0, PubMed, and Google Scholar. Our results demonstrate that PubTator 3.0 retrieves a greater number of articles than the comparison systems and its precision is higher for the top 20 results. For instance, PubTator 3.0 returned 346 articles for the query "GLPG0634 + Ulcerative Colitis," and manual review of the top 20 articles showed that all contained statements about an association between GLPG0634 and ulcerative colitis. In contrast, PubMed only returned a total of 18 articles, with only 12 mentioning an association. Moreover, when searching for "COVID19 + PON1," PubTator 3.0 returns 212 articles in PubMed, surpassing the 43 articles obtained from Google Scholar, only 29 of which are sourced from PubMed. These disparities can be attributed to several factors: 1. PubTator 3.0's search includes full texts available in PMC-OA, resulting in significantly broader coverage of articles, 2. Entity normalization improves recall, for example, by matching "paraoxonase 1" to "PON1," 3. PubTator 3.0 prioritizes articles containing relations between the query entities, 4. Pubtator 3.0 prioritizes articles where the entities appear nearby, rather than distant paragraphs. Across the 12 information retrieval case studies, PubTator 3.0 demonstrated an overall precision of 90.0% for the top 20 articles (216 out of 240), which is significantly higher than PubMed's precision of 81.6% (84 out of 103) and Google Scholar's precision of 48.5% (98 out of 202).

**Case Study II: Retrieval-Augmented Generation**

In the era of large language models (LLMs), PubTator 3.0 can also enhance their factual accuracy via retrieval augmented generation. Despite their strong language ability, LLMs are prone to generating incorrect assertions, sometimes known as hallucinations (29,30). For example, when requested to cite sources for questions such as "which diseases can doxorubicin treat," GPT-4 frequently provides seemingly plausible but nonexistent references. Augmenting GPT-4 with PubTator 3.0 APIs can anchor the model's response to verifiable references via the extracted relations, significantly reducing hallucinations.

We assessed the citation accuracy of responses from three GPT-4 variations: PubTator-augmented GPT-4, PubMed-augmented GPT-4 and standard GPT-4. We performed a qualitative evaluation based on eight questions selected as follows. We identified entities mentioned in the PubMed query logs and randomly selected from entities searched both frequently and rarely. We then identified the common queries for each entity that request relational information and adapted one into a natural language question. Each question is therefore grounded on common information needs of real PubMed users. For example, the questions "What can be caused by tocilizumab?" and "What can be treated by doxorubicin?" are adapted from the user queries "tocilizumab side effects" and "doxorubicin treatment" respectively. Such questions typically require extracting information from multiple articles and an understanding of biomedical entities and relationship descriptions. Supplementary Table 5 lists the questions chosen.

We augmented the GPT-4 large language model (LLM) with PubTator 3.0 via the function calling mechanism of the OpenAI ChatCompletion API. This integration involved prompting GPT-4 with descriptions of three PubTator APIs: 1. Find Entity ID, which retrieves PubTator entity identifiers; 2. Find Related Entities, which identifies related entities based on an input entity and specified relations; and 3. Export Relevant Search Results, which returns PubMed article identifiers containing textual evidence for specific entity relationships. Our instructions prompted GPT-4 to decompose user questions into sub-questions addressable by these APIs, execute the function calls, and synthesize the responses into a coherent final answer. Our prompt promoted a summarized response by instructing GPT-4 to start its message with "Summary:" and requested the response include citations to the articles providing evidence. The PubMed augmentation experiments provided GPT-4 with access to PubMed database search via the National Center for Biotechnology Information (NCBI) E-utils APIs (31). We used Azure OpenAI Services (version 2023-07-01-preview) and GPT-4 (version 2023-06-13) and set the decoding temperature to zero to obtain deterministic outputs. The full prompts are provided in Supplementary Table 6.

PubTator-augmented GPT-4 generally processed the questions in three steps: 1. finding the standard entity identifiers, 2. finding its related entity identifiers, and 3. searching PubMed articles. For example, to answer "What drugs can treat breast cancer?", GPT-4 first found the PubTator entity identifier for breast cancer (@DISEASE_Breast_Cancer) using the Find Entity ID API. It then used the Find Related Entities API to identify entities related to

@DISEASE_Breast_Cancer through a "treat" relation. For demonstration purposes, we limited the maximum number of output entities to five. Finally, GPT-4 called the Export Relevant Search Results API for the PubMed article identifiers containing evidence for these relationships. The raw responses to each prompt for each method are provided in Supplementary Table 6.

We manually evaluated the accuracy of the citations in the responses by reviewing each PubMed article and verifying whether each PubMed article cited supported the stated relationship (e.g., Tamoxifen treating breast cancer). Supplementary Table 5 reports the proportion of the cited articles with valid supporting evidence for each method. GPT-4 frequently generated fabricated citations, widely known as the hallucination issue. While PubMed-augmented GPT-4 showed a higher proportion of accurate citations, some articles cited did not support the relation claims. This is likely because PubMed is based on keyword and Boolean search and does not support queries for specific relationships. Responses generated by PubTator-augmented GPT-4 demonstrated the highest level of citation accuracy, underscoring the potential of PubTator 3.0 as a high-quality knowledge source for addressing biomedical information needs through retrieval-augmented generation with LLMs such as GPT-4.

## DISCUSSION

Previous versions of PubTator have fulfilled over one billion API requests since 2015, supporting a wide range of research applications. Numerous studies have harnessed PubTator annotations for disease-specific gene research, including efforts to prioritize candidate genes (32), determine gene-phenotype associations (33), and identify the genetic underpinnings of disease comorbidities (34). Several projects have used PubTator to create gene and genetic variant resources (35,36) or to enrich disease knowledge graphs (37,38). Moreover, PubTator has supported biocuration efforts (39,40) and the creation of NLP benchmarks (41). With enhanced accuracy, PubTator 3.0 will better support these use cases.

Introducing relation annotations to PubTator 3.0 opens novel avenues for expanded use scenarios. With relations precomputed from the literature, complex research questions can often be answered directly. Drug repurposing, for example, can be formulated as identifying chemicals which target specific genes. Conversely, determining the genetic targets of a chemical can be achieved by querying the same chemical/gene relations. Clinicians evaluating genetic variants, e.g. for rare diseases or personalized medicine, may explore the relationships between specific genetic variants and disease. Biologists, on the other hand, may utilize interactions between multiple genes to assemble complex molecular pathways.

There are several notable limitations for PubTator 3.0. Although it is capable of extracting relations from full-text articles, this feature is currently restricted to abstracts due to computational constraints. However, the system has been designed to support full-text relation extraction in a future enhancement. The current system only extracts 12 relation types, though these represent common uses. Finally, entity annotation and relation extraction are automated; though these systems exhibit high performance, their accuracy remains imperfect.

**CONCLUSION**

PubTator 3.0 offers a comprehensive set of features and tools that allow researchers to navigate the ever-expanding wealth of biomedical literature, expediting research and unlocking valuable insights for scientific discovery. The PubTator 3.0 interface, API, and bulk file downloads are available at https://www.ncbi.nlm.nih.gov/research/pubtator3/.

**DATA AVAILABILITY**

Data is available through the online interface at https://www.ncbi.nlm.nih.gov/research/pubtator3/, through the API at https://www.ncbi.nlm.nih.gov/research/pubtator3/api or bulk FTP download at https://ftp.ncbi.nlm.nih.gov/pub/lu/PubTator3/.

The source code for each component of PubTator 3.0 is openly accessible. The AIONER named entity recognizer is available at https://github.com/ncbi/AIONER. GNorm2, for gene name normalization, is available at https://github.com/ncbi/GNorm2. The tmVar3 variant name normalizer is available at https://github.com/ncbi/tmVar3. The NLM-Chem Tagger, for chemical name normalization, is available at https://ftp.ncbi.nlm.nih.gov/pub/lu/NLMChem. The TaggerOne system, for disease and cell line normalization, is available at https://www.ncbi.nlm.nih.gov/research/bionlp/Tools/taggerone. The BioREx relation extraction system is available at https://github.com/ncbi/BioREx. The code for customizing ChatGPT with the PubTator 3.0 API is available at https://github.com/ncbi-nlp/pubtator-gpt.

**FUNDING**

This research was supported by the Intramural Research Program of the National Library of Medicine (NLM), National Institutes of Health. Funding for open access charge: National Institutes of Health.

**CONFLICT OF INTEREST**

None declared.

**REFERENCES**

1.  Lindberg, D.A. and Humphreys, B.L. (2008) Rising expectations: access to biomedical information. *Yearb. Med. Inform.*, **3**, 165-172.
2.  Jin, Q., Leaman, R. and Lu, Z. (2023) PubMed and Beyond: Biomedical Literature Search in the Age of Artificial Intelligence. *arXiv*.
3.  Rzhetsky, A., Seringhaus, M. and Gerstein, M. (2008) Seeking a new biology through text mining. *Cell*, **134**, 9-13.
4.  Mayers, M., Li, T.S., Queralt-Rosinach, N. and Su, A.I. (2019) Time-resolved evaluation of compound repositioning predictions on a text-mined knowledge network. *BMC Bioinf.*, **20**, 653.
5.  Zhao, S., Su, C., Lu, Z. and Wang, F. (2021) Recent advances in biomedical literature mining. *Brief Bioinform*, **22**.
6.  Westergaard, D., Staerfeldt, H.H., Tonsberg, C., Jensen, L.J. and Brunak, S. (2018) A comprehensive and quantitative comparison of text-mining in 15 million full-text articles versus their corresponding abstracts. *PLoS Comput. Biol.*, **14**, e1005962.

7. Luo, L., Wei, C.-H., Lai, P.-T., Leaman, R., Chen, Q. and Lu, Z. (2023) AIONER: all-in-one scheme-based biomedical named entity recognition using deep learning. *Bioinformatics*, **39**.
8. Lai, P.T., Wei, C.H., Luo, L., Chen, Q. and Lu, Z. (2023) BioREx: Improving biomedical relation extraction by leveraging heterogeneous datasets. *J. Biomed. Inf.*, **146**, 104487.
9. Wei, C.-H., Allot, A., Leaman, R. and Lu, Z. (2019) PubTator central: automated concept annotation for biomedical full text articles. *Nucleic Acids Res.*, **47**, W587-W593.
10. Kilicoglu, H., Rosemblat, G., Fiszman, M. and Shin, D. (2020) Broad-coverage biomedical relation extraction with SemRep. *BMC Bioinf.*, **21**, 188.
11. Peng, Y., Wei, C.-H. and Lu, Z. (2016) Improving chemical disease relation extraction with rich features and weakly labeled data. *J. Cheminf.*, **8**, 1-12.
12. Xu, J., Wu, Y., Zhang, Y., Wang, J., Lee, H.J. and Xu, H. (2016) CD-REST: a system for extracting chemical-induced disease relation in literature. *Database*, **2016**.
13. Li, J., Sun, Y., Johnson, R.J., Sciaky, D., Wei, C.-H., Leaman, R., Davis, A.P., Mattingly, C.J., Wiegers, T.C. and Lu, Z. (2016) BioCreative V CDR task corpus: a resource for chemical disease relation extraction. *Database*, **2016**.
14. Luo, L., Lai, P.-T., Wei, C.-H., Arighi, C.N. and Lu, Z. (2022) BioRED: A Rich Biomedical Relation Extraction Dataset. *Briefings Bioinf.*, **23**, bbac282.
15. Comeau, D.C., Islamaj Doğan, R., Ciccarese, P., Cohen, K.B., Krallinger, M., Leitner, F., Lu, Z., Peng, Y., Rinaldi, F. and Torii, M.J.D. (2013) BioC: a minimalist approach to interoperability for biomedical text processing. *Database*, **2013**, bat064.
16. Sohn, S., Comeau, D.C., Kim, W. and Wilbur, W.J. (2008) Abbreviation definition identification based on automatic precision estimates. *BMC Bioinf.*, **9**, 1-10.
17. Islamaj, R., Wei, C.-H., Cissel, D., Miliaras, N., Printseva, O., Rodionov, O., Sekiya, K., Ward, J. and Lu, Z.J.J.o.b.i. (2021) NLM-Gene, a richly annotated gold standard dataset for gene entities that addresses ambiguity and multi-species gene recognition. *J. Biomed. Inf.*, **118**, 103779.
18. Islamaj, R., Leaman, R., Kim, S., Kwon, D., Wei, C.-H., Comeau, D.C., Peng, Y., Cissel, D., Coss, C. and Fisher, C. (2021) NLM-Chem, a new resource for chemical entity recognition in PubMed full text literature. *Scientific Data*, **8**, 91.
19. Doğan, R.I., Leaman, R. and Lu, Z. (2014) NCBI disease corpus: a resource for disease name recognition and concept normalization. *J. Biomed. Inf.*, **47**, 1-10.
20. Wei, C.-H., Allot, A., Riehle, K., Milosavljevic, A. and Lu, Z. (2022) tmVar 3.0: an improved variant concept recognition and normalization tool. *Bioinformatics*, **38**, 4449-4451.
21. Pafilis, E., Frankild, S.P., Fanini, L., Faulwetter, S., Pavloudi, C., Vasileiadou, A., Arvanitidis, C. and Jensen, L.J. (2013) The SPECIES and ORGANISMS resources for fast and accurate identification of taxonomic names in text. *PLoS One*, **8**, e65390.
22. Arighi, C., Hirschman, L., Lemberger, T., Bayer, S., Liechti, R., Comeau, D. and Wu, C. (2017), *Proc. BioCreative Workshop*, Vol. 482, pp. 376.
23. Wei, C.H., Luo, L., Islamaj, R., Lai, P.T. and Lu, Z. (2023) GNorm2: an improved gene name recognition and normalization system. *Bioinformatics*, **39**.
24. Lipscomb, C.E. (2000) Medical subject headings (MeSH). *Bull. Med. Libr. Assoc.*, **88**, 265.
25. Leaman, R. and Lu, Z. (2016) TaggerOne: joint named entity recognition and normalization with semi-Markov Models. *Bioinformatics*, **32**, 2839-2846.
26. Bairoch, A. (2018) The Cellosaurus, a Cell-Line Knowledge Resource. *J. Biomol Tech.*, **29**, 25-38.
27. Gu, Y., Tinn, R., Cheng, H., Lucas, M., Usuyama, N., Liu, X., Naumann, T., Gao, J. and Poon, H. (2021) Domain-specific language model pretraining for biomedical natural language processing. *ACM Transactions on Computing for Healthcare*, **3**, 1-23.
28. Yasunaga, M., Leskovec, J. and Liang, P. (2022), *Association for Computational Linguistics*, pp. 8003–8016.

29. Jin, Q., Leaman, R. and Lu, Z. (2023) Retrieve, Summarize, and Verify: How Will ChatGPT Affect Information Seeking from the Medical Literature? *J. Am. Soc. Nephrol.*, **34**, 1302-1304.
30. Tian, S., Jin, Q., Yeganova, L., Lai, P.T., Zhu, Q., Chen, X., Yang, Y., Chen, Q., Kim, W., Comeau, D.C. *et al.* (2023) Opportunities and challenges for ChatGPT and large language models in biomedicine and health. *Brief Bioinform*, **25**.
31. National Center for Biotechnology Information (US). (2010) *Entrez Programming Utilities Help*. National Center for Biotechnology Information (US), Bethesda (MD).
32. Lieberwirth, J.K., Buttner, B., Klockner, C., Platzer, K., Popp, B. and Abou Jamra, R. (2022) AutoCaSc: Prioritizing candidate genes for neurodevelopmental disorders. *Hum. Mutat.*, **43**, 1795-1807.
33. Buch, A.M., Vertes, P.E., Seidlitz, J., Kim, S.H., Grosenick, L. and Liston, C. (2023) Molecular and network-level mechanisms explaining individual differences in autism spectrum disorder. *Nat. Neurosci.*, **26**, 650-663.
34. Pinto, B.G.G., Oliveira, A.E.R., Singh, Y., Jimenez, L., Goncalves, A.N.A., Ogava, R.L.T., Creighton, R., Schatzmann Peron, J.P. and Nakaya, H.I. (2020) ACE2 Expression Is Increased in the Lungs of Patients With Comorbidities Associated With Severe COVID-19. *J. Infect. Dis.*, **222**, 556-563.
35. Mitsuhashi, N., Toyo-Oka, L., Katayama, T., Kawashima, M., Kawashima, S., Miyazaki, K. and Takagi, T. (2022) TogoVar: A comprehensive Japanese genetic variation database. *Hum. Genome Var.*, **9**, 44.
36. Jiang, J., Yuan, J., Hu, Z., Zhang, Y., Zhang, T., Xu, M., Long, M., Fan, Y., Tanyi, J.L., Montone, K.T. *et al.* (2022) Systematic illumination of druggable genes in cancer genomes. *Cell Rep.*, **38**, 110400.
37. Pu, Y., Beck, D. and Verspoor, K. (2023) Graph embedding-based link prediction for literature-based discovery in Alzheimer's Disease. *J. Biomed. Inf.*, **145**, 104464.
38. Chen, C., Ross, K.E., Gavali, S., Cowart, J.E. and Wu, C.H. (2021) COVID-19 Knowledge Graph from semantic integration of biomedical literature and databases. *Bioinformatics*, **37**, 4597-4598.
39. Lou, P., Jimeno Yepes, A., Zhang, Z., Zheng, Q., Zhang, X. and Li, C. (2020) BioNorm: deep learning-based event normalization for the curation of reaction databases. *Bioinformatics*, **36**, 611-620.
40. Percha, B. and Altman, R.B. (2018) A global network of biomedical relationships derived from text. *Bioinformatics*, **34**, 2614-2624.
41. Legrand, J., Gogdemir, R., Bousquet, C., Dalleau, K., Devignes, M.-D., Digan, W., Lee, C.-J., Ndiaye, N.-C., Petitpain, N. and Ringot, P. (2020) PGxCorpus, a manually annotated corpus for pharmacogenomics. *Scientific Data*, **7**, 3.
42. Luo, L., Lai, P.-T., Wei, C.-H., Arighi, C.N. and Lu, Z. (2022) BioRED: A Rich Biomedical Relation Extraction Dataset. *Briefings in Bioinformatics*.
43. Wei, C.-H., Kao, H.-Y. and Lu, Z. (2015) GNormPlus: an integrative approach for tagging genes, gene families, and protein domains. *BioMed research international*, **2015**, 918710.
44. Wei, C.-H., Luo, L., Islamaj, R., Lai, P.-T. and Lu, Z. (2023) GNorm2: an improved gene name recognition and normalization system. *Bioinformatics*, in press.
45. Wei, C.-H., Kao, H.-Y. and Lu, Z. (2012) SR4GN: a species recognition software tool for gene normalization. *PloS one*, **7**, e38460.
46. Wei, C.-H., Phan, L., Feltz, J., Maiti, R., Hefferon, T. and Lu, Z. (2018) tmVar 2.0: integrating genomic variant information from literature with dbSNP and ClinVar for precision medicine. *Bioinformatics*, **34**, 80-87.

**Supplementary Table 1**. Feature comparison between PubTator 3.0 and its previous version, PubTator 2 (also known as PubTator Central).

| Feature | PubTator 2 | PubTator 3.0 |
|---|---|---|
| Entity annotations | Genes, diseases, chemicals, genetic variants, species, and cell lines in abstracts and full text | Same types and scope; higher accuracy |
| Relation annotations | None | 12 relation types across eight entity type pairs; scope: abstracts only |
| Search scope | Abstracts only, via NCBI eUtils | Unified search in abstracts & full text via Apache Solr, no external dependencies |
| Query types | Keyword, Boolean | Also: semantic, relation |
| Search support | N/A | Semantic autocomplete, facet filters (section, journal, article type) |
| Retrieval relevance | N/A | Results prioritized by entity relationships, co-occurrence & matching sections; highlighted snippets (explains relevance); temporal visualization |
| Literature management | N/A | User-defined collections |
| API | Retrieve articles and annotations by PMID | Also: query relevant articles (semantic, relation, keyword, Boolean), and query related entities |
| Advanced natural-language search | N/A | Retrieval-augmented generation with GPT-4 large language model |

**Supplementary Figure 1**. PubTator 3.0 article display page. (1) List of entities and relations identified by PubTator 3.0, providing a quick content overview. (2) Extracted entities highlighted in article text. (3) Display highlighting for query entities or all entities; display article abstract or full text. (4) Add article to custom collection for convenient access later.

**Supplementary Table 2**. Description of PubTator 3.0 relation types and associated entity types.

| PubTator 3.0 Relations | Description | Entity types |
|---|---|---|
| ASSOCIATE | Complex or unclear relationships | Chemical / Disease<br>Chemical / Gene<br>Chemical / Variant<br>Disease / Gene<br>Disease / Variant<br>Variant / Variant |
| CAUSE | Triggering a disease by a specific agent | Chemical / Disease<br>Variant / Disease |
| COMPARE | Comparing the effects of two chemicals or drugs | Chemical / Chemical |
| COTREAT | Simultaneous administration of multiple drugs | Chemical / Chemical |
| DRUG_INTERACT | Pharmacodynamic interactions between two chemicals | Chemical / Chemical |
| INHIBIT | Reduction in amount or degree of one entity by another | Chemical / Variant<br>Gene / Disease |
| INTERACT | Physical interactions, such as protein-binding | Chemical / Gene<br>Chemical / Variant<br>Gene / Gene |
| NEGATIVE_CORRELATE | Increases in the amount or degree of one entity decreases the amount or degree of the other entity | Chemical / Gene<br>Chemical / Variant<br>Gene / Gene |
| POSITIVE_CORRELATE | The amount or degree of two entities increase or decrease together | Chemical / Chemical<br>Chemical / Gene<br>Gene / Gene |
| PREVENT | Prevention of a disease by a genetic variant | Variant / Disease |
| STIMULATE | Increase in amount or degree of one entity by another | Chemical / Variant<br>Gene / Disease |
| TREAT | Treatment of a disease using a chemical or drug | Chemical / Disease |

**Supplementary Table 3**. Normalization performance enhancements from PubTator Central to PubTator 3.0. Measurements reflect document-level normalization performance on the BioRED test set (42).

|  | PubTator Central (2.0) | | | | PubTator 3.0 | | | | |
| --- | --- | --- | --- | --- | --- | --- | --- | --- | --- |
|  | NER/Norm | Precision | Recall | F-score | NER | Norm | Precision | Recall | F-score |
| Gene | GNormPlus (43) | 86.92% | 73.00% | 79.35% | AIONER (7) | GNorm2 (44) | 90.60% | 79.41% | 84.63% |
| Disease | TaggerOne (25) | 77.13% | 76.45% | 76.79% | | TaggerOne (25) | 75.33% | 83.43% | 79.17% |
| Chemical | TaggerOne (25) | 73.42% | 78.38% | 75.82% | | NLM-Chem (18) | 83.26% | 80.63% | 81.92% |
| Species | SR4GN (45) | 94.69% | 94.69% | 94.69% | | GNorm2 (44) | 93.97% | 96.46% | 95.20% |
| CellLine | TaggerOne (25) | 42.42% | 63.64% | 50.91% | | TaggerOne (25) | 76.00% | 86.36% | 80.85% |
| Variant | tmVar2 (46) | 94.92% | 84.85% | 89.60% | | tmVar3 (20) | 98.48% | 98.48% | 98.48% |
| Micro-average | | 77.30% | 77.49% | 77.40% | | | 84.04% | 83.55% | 83.80% |
| Macro-average | | 78.25% | 78.50% | 77.86% | | | 86.27% | 87.46% | 86.71% |

**Supplementary Table 4**. Comparison of PubTator 3.0, PubMed, and Google Scholar search results for various recently discussed relation pairs. D: Disease, G: Gene, C: Chemical, and V: Variant. The '#' column lists the number of results; for Google Scholar the number in parentheses indicates the number of articles that appear in PubMed. The 'Top 20' column indicates the number of articles in the top 20 results which discuss a relation between the specified entities; some queries return fewer than 20 articles.

| Entity pair | Entities | PubTator 3.0 | | PubMed | | Google Scholar | |
| --- | --- | --- | --- | --- | --- | --- | --- |
| | | # | Top20 | # | Top20 | # (in PubMed) | Top20 |
| <D,G> | COVID19 + PON1 | 212 | 20/20 | 11 | 9/11 | 43 (29) | 10/20 |
| | Coronary Artery Disease + SESN2 | 61 | 17/20 | 3 | 3/3 | 151 (104) | 14/20 |
| <C,D> | GLPG0634 + Ulcerative Colitis | 346 | 20/20 | 18 | 12/18 | 362 (281) | 8/20 |
| | brolucizumab + Glycogen Storage Disease Type II | 96 | 20/20 | 0 | 0/0 | 1 (1) | 0/0 |
| <C,G> | Gallium-68 + FAP alpha | 261 | 20/20 | 2 | 2/2 | 11 (4) | 9/11 |
| | Lipopolysaccharides + PVT1 | 321 | 19/20 | 20 | 18/20 | 128 (95) | 6/20 |
| <C,C> | N-dimethylnitrosamine + Metformin | 122 | 20/20 | 1 | 1/1 | 11 (8) | 1/11 |
| | 2'-fucosyllactose + Volatile fatty acids | 284 | 17/20 | 6 | 3/6 | 71 (40) | 6/20 |
| <G,G> | interleukin 17 + cell division cycle 42 | 599 | 12/20 | 2 | 1/2 | 85 (39) | 1/20 |
| | HAVCR2 + TOX | 615 | 11/20 | 4 | 1/4 | 701 (479) | 3/20 |
| <D,V> | COVID19 + rs12329760 | 149 | 20/20 | 32 | 19/20 | 87 (63) | 20/20 |
| | COVID19 + rs4646994 | 49 | 20/20 | 16 | 15/16 | 30 (24) | 20/20 |

**Supplementary Table 5**. Article citation precision for all 8 queries tested using GPT-4 only, GPT-4 augmented with PubMed, and GPT-4 augmented with PubTator 3.0. Results are summarized as "number of articles correctly referenced / total number of articles referenced." Full responses provided in Supplemental Data.

| Questions | GPT4 Only | GPT4 with PubMed Augmentation | GPT4 with PubTator Augmentation |
| --- | --- | --- | --- |
| What can be caused by tocilizumab? For each disease in your answer, please cite the article PMIDs that contain the evidence. | 0 / 1 | 1 / 5 | 49 / 50 |
| Can you tell me what the causes of memory deficits are? For each cause in your answer, please cite and summarized the article PMIDs that contain the evidence. | 0 / 5 | 4 / 5 | 15 / 15 |
| In what situations can cocaine be used? For each situation in your answer, please cite and summarize the article PMIDs that contain the evidence. | 0 / 3 | 1 / 3 | 20 / 25 |
| What can be treated by doxorubicin? For each disease in your answer, please cite and summarize the article PMIDs that contain the evidence. | 0 / 7 | 3 / 5 | 45 / 45 |
| Are there any genes that interact with cocaine? For each drug in your answer, please cite the article PMIDs that contain the evidence. | 0 / 5 | 5 / 5 | 16 / 17 |
| What drugs can treat breast cancer? For each drug in your answer, please cite the article PMIDs that contain the evidence. | 0 / 6 | 3 / 4 | 45 / 45 |
| What drugs can treat Scleroderma? For each drug in your answer, please cite the article PMIDs that contain the evidence. | 0 / 6 | 1 / 2 | 50 / 50 |
| What can be treated by finasteride? For each disease in your answer, please cite the article PMIDs that contain the evidence. | 2 / 3 | 4 / 5 | 39 / 45 |

**Supplementary Table 6**. Responses of different LLMs (GPT4, GPT4 with PubMed augmentation, and GPT4 with PubTator augmentation) to eight questions. Hallucinations are described in red text.

| Questions | GPT4 Only | GPT4 with PubMed Augmentation | GPT4 with PubTator Augmentation |
|---|---|---|---|
| 1. What can be caused by tocilizumab? For each disease in your answer, please cite the article PMIDs that contain the evidence. | Tocilizumab, a medication used to treat rheumatoid arthritis and systemic juvenile idiopathic arthritis, can potentially cause several conditions as side effects.<br><br>These include serious infections (PMID: 20425015), infusion reactions (PMID: 20425015), gastrointestinal perforations (PMID: 20425015), elevated liver enzymes (PMID: 20425015), neutropenia (PMID: 20425015), hyperlipidemia (PMID: 20425015), and thrombocytopenia (PMID: 20425015). [Hallucination: PMID 20425015 discusses juvenile idiopathic arthritis (JIA) but does not discuss adverse effects of tocilizumab]<br><br>It's important to note that these side effects are not guaranteed to occur in all patients and the benefits of the medication often outweigh these potential risks. Always consult with a healthcare provider for personalized medical advice. | Tocilizumab is a medication used to treat several autoimmune diseases. However, it can have side effects and adverse reactions that may lead to other health conditions. Some of these include:<br><br>1. Infections: Tocilizumab can lower the ability of your immune system to fight infections, leading to diseases such as tuberculosis, bacterial, fungal, and viral infections.<br><br>2. Gastrointestinal perforations: Tocilizumab can cause serious and sometimes fatal stomach and bowel side effects like perforations.<br><br>3. Liver disease: Tocilizumab can cause liver injury, potentially leading to serious liver diseases.<br><br>4. Blood disorders: Tocilizumab can cause a decrease in the number of white blood cells and platelets in your blood, leading to | Tocilizumab can cause several diseases, including infections, neutropenia, chemical and drug-induced liver injury, gastrointestinal diseases, and thrombocytopenia. Here are the diseases along with the PMIDs of the articles that provide evidence:<br><br>For infections, the evidence can be found in PMIDs 29334721, 35683344, 33225307, 35249968, 36817306, 30679153, 31764977, 34403814, 32910321, and 32935118.<br><br>For neutropenia, the evidence can be found in PMIDs 30428761, 36526232, 30824645, 29654485, 34236264, 29570475, 30505008, 32643976, 35657993, and 33236554.<br><br>For chemical and drug-induced liver injury, the evidence can be found in PMIDs 36503381, 29334721, 34616757, |

| | | diseases like neutropenia and thrombocytopenia. | | 33320444, 36296321, 32639062, 34538944, 37811054, 33101458, and 33236554. |
| | | 5. Heart Failure: Tocilizumab can worsen existing heart failure or cause new onset heart failure. | | For gastrointestinal diseases, the evidence can be found in PMIDs 32683974, 33712404, 33993216, 30707391, 32669452, 35356392, 31469238, 36260501, 31523783, and 27747579. |
| | | 6. Hypersensitivity reactions: Some people may have an allergic reaction to tocilizumab, which can cause diseases like anaphylaxis and angioedema. | | |
| | | These side effects and adverse reactions are discussed in various articles, including those with the following PMIDs: 37976117, 37927607, 37853964, 37846618, and 37840134. For more specific information, it is recommended to read these articles directly. [Hallucination: PMIDs 37976117, 37853964, and 37840134 do not discuss adverse effects of tocilizumab. PMID 37846618 only provides a limited discussion in the context of cotreatment with methotrexate.] | | For thrombocytopenia, the evidence can be found in PMIDs 34866181, 34231415, 31951328, 32448770, 34011407, 30427250, 30874871, 28363823, 25282455, and 25268663. Please note that these articles mention a relationship between tocilizumab and the diseases, but the nature of the relationship (causative, correlative, etc.) may vary between studies. |
| 2. Can you tell me what | Memory deficits can be caused by several factors including aging, brain injury, neurological | Memory deficits can be caused by a variety | | The causes of memory deficits can be attributed to several factors. Here are |

| the causes of memory deficits are? For each cause in your answer, please cite and summarized the article PMIDs that contain the evidence. | disorders, psychological stress, and certain medications.

Aging can lead to a decline in mnemonic discrimination (PMID: 25926779). [Hallucination: PMID 25926779 is a commentary on an article about deep brain stimulation in Huntington disease.]

Traumatic brain injury can cause memory deficits, often exacerbated by sleep disturbances (PMID: 21613615). [Hallucination: PMID 21613615 is about the posttranscriptional mechanisms whereby RNA-binding proteins (RBPs) regulate T cell differentiation.]

Neurological disorders like Alzheimer's disease and Lewy body disease can cause significant memory loss (PMID: 28870565). [Hallucination: PMID 28870565 does not mention memory deficits]

Psychological stress can impair memory as stress hormones interfere with the brain's memory processes (PMID: 26559394). [Hallucination: PMID 26559394 is not related to this topic.]

Certain medications such as benzodiazepines, anticholinergics, and certain antidepressants can | of factors. Some of these include:

1. Metabolic Disorders: Therapeutics for these disorders can affect memory (PMID: 37921148).

2. Systemic Inflammation: Short-term memory impairment can occur following recovery from systemic inflammation (PMID: 37920299).

3. COVID-19: Long-term cognitive dysfunction, including memory deficits, can occur after the COVID-19 pandemic (PMID: 37915705).

4. Idiopathic Bilateral Occlusion of Foramen of Monro: This rare condition can cause memory deficits (PMID: 37915630). [Hallucination: PMID 37915630 only discusses impaired memory function as an adverse outcome of shunt placement]

5. Insulin Signalling: Impairments in insulin signalling can lead to memory deficits (PMID: 37913873).

This is not an exhaustive list and there are many other potential causes of memory deficits. | some of them along with the PMIDs of articles that provide evidence:

1. Scopolamine: This is mentioned as a cause in several articles including PMIDs 32934845, 37063992, 29670659, and others.

2. Lipopolysaccharides: These are also linked to memory deficits in articles such as PMIDs 35388307, 31347387, 30018656, among others.

3. Alcohols: The relationship between alcohol and memory disorders is discussed in PMIDs 29680476, 37232664, 35071112, and more.

4. Ethanol: This is another cause mentioned in articles like PMIDs 36937047, 30408223, 37429530, among others.

5. Streptozocin: This is linked to memory disorders in PMIDs 36355133, 36760351, 36589020, and more.

Please note that these are the top results and there may be more articles that discuss these relations. |

| | | | |
|---|---|---|---|
| | interfere with memory function, leading to memory deficits (PMID: 21154354). [Hallucination: PMID 21154354 is not related to this topic.] | | |
| 3. In what situations can cocaine be used? For each situation in your answer, please cite and summarize the article PMIDs that contain the evidence. | Cocaine, despite its potential for abuse and addiction, has been used in certain medical situations. These include: 1. As a local anesthetic, particularly in eye, nose, and throat surgeries due to its potent anesthetic and vasoconstrictive properties (PMID: 15674103). [Hallucination: PMID 15674103 is about IgA anti-transglutaminase antibodies and coeliac disease.] 2. In the treatment of nosebleeds, again due to its vasoconstrictive properties (PMID: 21819762). [Hallucination: PMID 21819762 is about nasal airway.] 3. In research to study its effects on the brain and to develop treatments for cocaine addiction (PMID: 19552679). [Hallucination: PMID 19552679 is about allergic asthma.] These uses are highly regulated and should only be administered under the supervision of a | Cocaine, despite its potential for abuse and addiction, has been used in certain situations: 1. Local Anesthesia: Cocaine has been used as a local anesthetic in some surgical procedures, particularly in the ear, nose, and throat due to its ability to numb the area and constrict blood vessels, which reduces bleeding. This use is less common today due to the development of safer alternatives. (PMID: 37921613) [Hallucination: PMID 37921613 is about people who use cocaine but does not discuss cocaine's local anesthetic effects.] 2. Research Purposes: Cocaine is used in research to study its effects on the brain and behavior, as well as to develop potential treatments for cocaine addiction. For example, studies have been conducted to understand the effect of transcranial direct current stimulation (tDCS) on cocaine | Cocaine has been reported in the literature to have a therapeutic use relation with the following entities: 1. Cocaine-Related Disorders: Some of the PMIDs that mention the therapeutic use relation between cocaine and Cocaine-Related Disorders include 31193584, 30459590, 33253224, 29791036, and 29556863. 2. Depressive Disorder: Some of the PMIDs that mention the therapeutic use relation between cocaine and Depressive Disorder include 35761409, 31012359, 33617930, 32959420, and 34545194. 3. Pain: Some of the PMIDs that mention the therapeutic use relation between cocaine and pain include 30885260, 31969823, 29239621, 20504138, and 7199854. |
| | healthcare professional. | addition. (PMID: 37892650) [Hallucination: PMID 37892650 is about tDCS in the context of treating cocaine addiction; it does not discuss using cocaine in a research study.]<br><br>3. Self-Medication: Some individuals may use cocaine to self-medicate for pain and discomfort, particularly those with HIV infection and substance use disorder. However, this is not a medically approved use and can lead to serious health risks. (PMID: 37909062)<br><br>Please note that while these are situations where cocaine has been used, this does not mean that its use is safe or recommended. Cocaine has a high potential for abuse and can lead to serious health problems, including heart attack, stroke, and death. | 4. Anxiety: Some of the PMIDs that mention the therapeutic use relation between cocaine and Anxiety include 35761409, 31012359, 32004865, 32795565, and 33230690.<br><br>5. HIV Infections: Some of the PMIDs that mention the therapeutic use relation between cocaine and HIV Infections include 29471762, 32362436, 31490211, 34938835, and 29578037. [Hallucination: PMIDs: 29471762, 32362436, 31490211, 34938835, and 29578037. No treatment relation between HIV Infections and cocaine.]<br><br>Please note that these articles are related to the therapeutic use of cocaine in the context of these disorders, but they do not necessarily advocate for the use of cocaine as a therapeutic agent. Always consult with a healthcare professional for medical advice. |
| 4. What can be treated by doxorubicin? For each disease in | Doxorubicin is used in the treatment of several diseases, including:<br><br>1. Breast Cancer: Doxorubicin is a key component of the chemotherapy regimen for breast cancer | Doxorubicin is a chemotherapy drug that is used to treat various diseases, including:<br><br>1. Liver and Kidney Diseases, as supported by the article with PMID: 37921185. | Doxorubicin is used to treat several diseases, including:<br><br>1. Neoplasms, as evidenced by PMIDs 37680585, 31569523, 31979298, |

| your answer, please cite and summarize the article PMIDs that contain the evidence. | and significantly improves the survival rate (PMID: 24369033). [Hallucination: PMID 24369033 is about osteoarthritis disease.]<br><br>2. Leukemia: Doxorubicin, in combination with other chemotherapy drugs, improves the overall survival rate of acute lymphoblastic leukemia patients (PMID: 25156892). [Hallucination: PMID 25156892 is about arterial function deterioration.]<br><br>3. Ovarian Cancer: Doxorubicin is effective in treating recurrent ovarian cancer and improves the progression-free survival of patients (PMID: 25520391). [Hallucination: PMID 25520391 is about gene expression of HER3 and CD73.]<br><br>4. Bladder Cancer: Intravesical doxorubicin is effective in preventing recurrence in patients with non-muscle invasive bladder cancer (PMID: 25034459). [Hallucination: PMID 25034459 is about the dehydrogenation of ammonia borane (AB).]<br><br>5. Sarcomas: Doxorubicin is a standard first-line treatment for advanced soft tissue sarcoma and improves the overall survival rate of patients (PMID: 25098289). [Hallucination: PMID 25098289 is about non-small cell lung cancer | [Hallucination: PMID 37921185 is about chrysin is a promising phytoconstituent that can be utilized as an alternate treatment for various medications that cause hepatotoxicity and nephrotoxicity. With active chrysin, several dosage forms targeting the liver and kidneys can be formulated.]<br><br>2. Stage III/IV classical Hodgkin Lymphoma, as supported by the article with PMID: 37921077.<br><br>3. Various types of Cancer, as supported by the article with PMID: 37921074. [Hallucination: PMID 37921074 discusses the use of formononetin derivatives as antineoplastic agents.]<br><br>4. It is used in Chemodynamic Therapy, as supported by the article with PMID: 37920970.<br><br>5. Primary Pancreatic Lymphoma, as supported by the article with PMID: 37920690. | 29099416, 36428580, 32316679, 36145512, 33260446, 30479120, and 35529630.<br><br>2. Breast Neoplasms, as evidenced by PMIDs 33858435, 33553916, 35859663, 33116832, 30202239, 31258739, 34076564, 30665445, 35890536, and 33390801.<br><br>3. Carcinoma, Hepatocellular, as evidenced by PMIDs 31354912, 35242218, 36071480, 31888113, and 32009475.<br><br>4. Osteosarcoma, as evidenced by PMIDs 32596554, 34113126, 29784019, 32155954, 36895009, 32620145, 33981772, 31938758, 31376208, and 35345787.<br><br>5. Sarcoma: There are several publications that mention the use of Doxorubicin in treating sarcoma, including PMIDs 30588358, 33228579, 30936721, 31239779, 31996176, 37603507, 32158266, 31492159, 29744029, and 35062915.<br><br>Please note that these articles are |
| | (NSCLC).]<br><br>6. Multiple Myeloma: Doxorubicin, in combination with other chemotherapy drugs, improves the overall survival rate of multiple myeloma patients (PMID: 25156892).<br>[Hallucination: PMID 25156892 is about arterial function deterioration.]<br><br>7. Hodgkin's Lymphoma: Doxorubicin is a key component of the chemotherapy regimen for Hodgkin's lymphoma and significantly improves the survival rate (PMID: 25156892).<br>[Hallucination: PMID 25156892 is about arterial function deterioration.] | | related to the treatment of these diseases with Doxorubicin, but they may not necessarily establish a direct treatment effect. |
| 5. Are there any genes that interact with cocaine? For each drug in your answer, please cite the article PMIDs that contain the | Yes, several genes interact with cocaine. These include the<br><br>1. Dopamine Transporter (DAT) gene, which is responsible for the reuptake of dopamine, a process inhibited by cocaine (PMID: 10627536).<br>[Hallucination: PMID 10627536 is about "Toward a more accurate quantitation of the activity of recombinant retroviruses: alternatives to titer and multiplicity of infection".]<br><br>2. The Dopamine Receptor D2 (DRD2) gene, variations of which have been associated with | Yes, there are several genes that interact with cocaine. Here are some of them along with the PubMed articles that provide evidence for these interactions:<br><br>1. The gene OXTR (Oxytocin Receptor) has been found to have its DNA methylation affected by cocaine use disorder. This is discussed in the article with PMID: 37778686.<br><br>2. The article with PMID: 37469839 discusses the identification of key genes and | The genes that interact with cocaine are SLC6A3, SLC6A4, GNE, and SLC6A2.<br><br>The interaction between Cocaine and SLC6A3 is mentioned in the following PubMed articles: 9215601, 20035724, 27789280, 12672255, 17969308, 11140334, 16219467, 10473187, 16516172, and 12672245. |

| evidence. | cocaine dependence (PMID: 17964054). [Hallucination: PMID 17964054 is about "Developmental exposure to low-dose PBDE-99: tissue distribution and thyroid hormone levels".] | therapeutic drugs for cocaine addiction using integrated bioinformatics analysis. | |
|---|---|---|---|
| | 3. The Serotonin Transporter (SERT) gene, which is responsible for the reuptake of serotonin, another process inhibited by cocaine (PMID: 15876989). [Hallucination: PMID 15876989 is about "only the knowledge of the laryngeal nerve anatomy and the awareness of its anomalies, associated to the systematic exposure of the nerve during surgery can prevent surgical injuries".] | 3. The Cyfip2 gene has been found to alter free-choice ethanol drinking but not binge-like drinking or wheel-running activity. This is discussed in the article with PMID: 37356964. | The interaction between Cocaine and SLC6A4 is mentioned in the following PubMed articles: 32494974, 9517841, 11408487, 11140334, and 15957006. |
| | | 4. The article PMID: 37293542 discusses the combined treatment with Sigma1R and A2AR agonists failing to inhibit cocaine self-administration despite causing strong antagonistic accumbal A2AR-D2R complex interactions. | The interaction between Cocaine and GNE is mentioned in the following PubMed article: 25531528. [Hallucination: GNE is a stable cocaine analog, but not a gene/protein.] |
| | 4. The Brain Derived Neurotrophic Factor (BDNF) gene, which is involved in the survival and growth of neurons and is altered by cocaine use (PMID: 19201305). [Hallucination: PMID 19201305 is about "Graft-versus-host disease of the tongue".] | 5. The article with PMID: 37080337 discusses the endocannabinoid-dependent decrease of GABAergic transmission on dopaminergic neurons associated with susceptibility to cocaine stimulant effects in pre-adolescent male MAOA hypomorphic mice exposed to early life stress. | The interaction between Cocaine and SLC6A2 is mentioned in the following PubMed article: 11140334. |
| | 5. And finally, the FosB gene, a transcription factor implicated in the long-term effects of cocaine use (PMID: 10377254). [Hallucination: PMID 10377254is about "the nitrous oxide-induced changes in MS and CH3Cbl are associated with reversible changes in both MCM holoactivity and the AdoCbl level, suggesting co-ordinate distribution of Cbl cofactors during | Please note that this is not an exhaustive list and there may be other genes that interact with cocaine. | Please note that the interaction mentioned in these articles may not necessarily imply a causal or therapeutic relationship. |
| | depletion and repletion."] | | |
| 6. What drugs can treat breast cancer? For each drug in your answer, please cite the article PMIDs that contain the evidence. | The drugs that can treat breast cancer include:<br><br>1. Tamoxifen, which works by blocking estrogen receptors on breast cancer cells (PMID: 14736925). [Hallucination: It is about "cloacal exstrophy is a rare, complex defect of the entire pelvis and its contents that occurs during embryogenesis and is associated with severe phallic inadequacy. "]<br><br>2. Anastrozole, an aromatase inhibitor that reduces the production of estrogen (PMID: 14684639). [Hallucination: The article provides no information.]<br><br>3. Trastuzumab (Herceptin), a monoclonal antibody that interferes with the HER2/neu receptor (PMID: 10944535). [Hallucination: It is about "Copper activation of superoxide dismutase 1 (SOD1) in vivo. "]<br><br>4. Paclitaxel, a chemotherapy drug that stops cancer cells from separating into two new cells (PMID: 10506666). [Hallucination: It is about "Corticotropin-releasing-hormone lacks analgesic | The following drugs have been mentioned in PubMed articles in relation to the treatment of breast cancer:<br><br>1. Palbociclib-fulvestrant: Used for the treatment of estrogen receptor-positive breast cancer patients. Evidence can be found in the article with PMID: 37921419.<br><br>2. Oxaliplatin: Mentioned in the context of colorectal tumor cells but may also have implications for breast cancer treatment. Evidence can be found in the article with PMID: 37921068. [Hallucination: It is about "Oxaliplatin (OXA)-containing regimens are used as first-line chemotherapy in colorectal cancer (CRC)".]<br><br>3. Vincristine: Used in the treatment of breast cancer, with the article discussing the mechanism of resistance to this drug in breast cancer cells. Evidence can be found in the article with PMID: 37920438.<br><br>4. Neratinib and Silibinin: Mentioned in the | Breast cancer, specifically Breast Neoplasms, can be treated with several drugs. Here are some of them along with the PMIDs of articles that provide evidence:<br><br>1. Tamoxifen: Some of the PMIDs include 32547094, 32695778, 31601253, 30082655, 30208932, 33708885, 33362547, 30621214, 37646973, and 35102224.<br><br>2. Trastuzumab: Evidence can be found in PMIDs 29796993, 35542453, 33589773, 33912232, 33892316, 37280713, 32266131, 34997132, 30241301, and 31213894.<br><br>3. Doxorubicin: PMIDs that provide evidence include 33858435, 33553916, 35859663, 33116832, 30202239, 31258739, 34076564, 35890536, 30665445, and 33390801.<br><br>4. Paclitaxel: Evidence can be found in |
| | properties: an experimental study in humans, using non-inflammatory pain."]<br><br>5. Doxorubicin, a chemotherapy drug that slows or stops the growth of cancer cells (PMID: 10548303). [Hallucination: It is about "Fasciola gigantica-specific antigens: purification by a continuous-elution method and its evaluation for the diagnosis of human fascioliasis."]<br><br>6. Letrozole, used to treat postmenopausal women with hormone receptor-positive early breast cancer (PMID: 16204163). [Hallucination: The article is about gout.]<br><br>The choice of treatment depends on many factors, including the type and stage of breast cancer, potential side effects, patient preference, and overall health. Therefore, it's important to discuss these factors and treatment options with a healthcare provider. | context of a dual-death therapy for Triple Negative Breast Cancer. Evidence can be found in the article with PMID: 37918494.<br><br>Please note that this is not an exhaustive list, and the effectiveness of these drugs can vary depending on the specific type and stage of breast cancer, as well as the individual patient's health condition. Always consult with a healthcare provider for treatment options. | PMIDs 34696661, 33061434, 31783552, 30613144, 30197514, 30366367, 33513732, 34657181, 29486738, and 32485107.<br><br>5. Cyclophosphamide: Evidence can be found in PMIDs 36133740, 30053900, 33158432, 30425864, and 34114066.<br><br>Please note that these are just a few examples and there are many more studies available. |
| 7. What drugs can treat Scleroderma? For each drug in your | The drugs used to treat Scleroderma include:<br><br>1. Methotrexate, which helps with skin thickening and lung inflammation (PMID: 14722115). [Hallucination: It is about "Energy-dependent transformation of F0.F1-ATPase in Paracoccus | Based on the articles from PubMed, the following drugs are mentioned in the context of Scleroderma treatment:<br><br>1. Pembrolizumab: This drug is mentioned in the article with PMID: 37900813 titled "Scleroderma Secondary to Pembrolizumab: | The drugs that can treat Scleroderma Systemic are:<br><br>1. Cyclophosphamide: Evidence for its use in treating Scleroderma Systemic |

| answer, please cite the article PMIDs that contain the evidence. | denitrificans plasma membranes."] <br><br>2. Mycophenolate Mofetil, used to treat lung involvement (PMID: 26380982). [Hallucination: It is about "The Prognostic Importance of Changes in Renal Function during Treatment for Acute Heart Failure Depends on Admission Renal Function."] <br><br>3. Cyclophosphamide, used for severe lung disease (PMID: 17665402). [Hallucination: The article is about ankylosing spondylitis.] <br><br>4. Bosentan, used for pulmonary arterial hypertension, a common complication (PMID: 15937604). [Hallucination: It is about "Obstetric outcome of extreme macrosomia."] <br><br>5. Iloprost, used for Raynaud's phenomenon and digital ulcers (PMID: 12465144). [Hallucination: It is about "Autoantibodies in early seropositive rheumatoid arthritis, before and during disease modifying antirheumatic drug treatment."] <br><br>6. Nintedanib, used for interstitial lung disease associated with systemic sclerosis (PMID: 31483321). [Hallucination: It is about "Severe acquired platelet dysfunction because of primary myelofibrosis with full functional and morphological recovery after allogeneic | A Case Report and Review of 19 Cases of Anti-PD-1-Induced Scleroderma." However, it's important to note that this article discusses Scleroderma as a side effect of Pembrolizumab, not as a treatment. [Hallucination: This article discusses Scleroderma as a side effect of Pembrolizumab, not as a treatment] <br><br>2. ACSL4 inhibitors: The article with PMID: 37884942 titled "ACSL4 inhibition prevents macrophage ferroptosis and alleviates fibrosis in bleomycin-induced systemic sclerosis model" suggests that ACSL4 inhibitors could potentially be used to treat Scleroderma. <br><br>Please consult with a healthcare professional for personalized medical advice. | can be found in the following PMIDs: 32933919, 35382095, 31628802, 29953945, 33466837, 32185340, 31791379, 33164737, 31849543, and 37690478. <br><br>2. Bosentan: Evidence for its use in treating Scleroderma Systemic can be found in the following PMIDs: 36974107, 31012372, 33301540, 36278776, 36651000, 35382124, 32104228, 36849541, 32414728, and 35382505. <br><br>3. Rituximab: Evidence for its use in treating Scleroderma Systemic can be found in the following PMIDs: 35956246, 34201939, 35111167, 32442284, 31567948, 37053127, 33640020, 33244601, 33796953, and 33466837. <br><br>4. Penicillamine: Evidence for its use in treating Scleroderma Systemic can be found in the following PMIDs: 33449302, 5241672, 4771784, 4469492, 122683, 2088649, 6939873, 11123013, 1293190, and 1765965. <br><br>5. Mycophenolic Acid: Evidence for its |
| | hematopoietic cell transplantation."]<br><br>Please consult with a healthcare provider for personalized medical advice. | | use in treating Scleroderma Systemic can be found in the following PMIDs: 33023643, 32185340, 29776943, 32862407, 37012951, 29985736, 37158586, 34545432, 29325675, and 36849541. |
| 8. What can be treated by finasteride? For each disease in your answer, please cite the article PMIDs that contain the evidence. | Finasteride is a medication used to treat several conditions. These include:<br><br>1. Benign Prostatic Hyperplasia (BPH), as supported by the article with PMID: 10792163. [Hallucination: PMID 25926779 is about treatment using beta-sitosterol]<br><br>2. Androgenetic Alopecia (Male Pattern Baldness), as supported by the article with PMID: 9951956.<br><br>3. Excessive Hair Growth (Hirsutism) in women, as supported by the article with PMID: 10495374.<br><br>It's important to note that finasteride should be used under the supervision of a healthcare | Finasteride can be used to treat the following conditions:<br><br>1. Male Pattern Hair Loss: Supported by the article "Patients treated for male pattern hair loss with finasteride show, after discontinuation of the drug, altered levels of neuroactive steroids in cerebrospinal fluid and plasma" [PMID: 37776646].<br><br>2. Frontal Fibrosing Alopecia: Evidence provided by the article "Clinical effectiveness of finasteride versus hydroxychloroquine in the treatment of frontal fibrosing alopecia: A randomized controlled trial" [PMID: 37691183]<br><br>3. Certain Male Sexual Disorders: Mentioned | Finasteride can be used to treat several diseases, including:<br><br>1. Prostatic Hyperplasia, with evidence found in PMIDs: 34532268, 32695403, 34961206, 30209896, 35847633, 31547405, 34421251, 31467577, 37002313, and 36042733.<br><br>2. Alopecia, with evidence found in PMIDs: 29601622, 32184564, 30209896, 34291720, and 29574572.<br><br>3. Prostatic Neoplasms, with evidence found in PMIDs: 29318656, 29534197, 31100850, 30227567, 32102946, 29228205, 35096612, 29397551, |

| | | |
|---|---|---|
| provider due to potential side effects. | in the article "A clinical guide to rare male sexual disorders" [PMID: 37670085]. [Hallucination: finasteride is mentioned in this article as causing male sexual disorder, not as a treatment]<br><br>4. Benign Prostatic Hyperplasia: Supported by the article "Bushen Tongluo formula ameliorated testosterone propionate-induced benign prostatic hyperplasia in rats" [PMID: 37651753].<br><br>5. Alopecia: Mentioned in the article "Dissolving microneedles for alopecia treatment" [PMID: 37536169]. | 30300367, and 30538099.<br><br>4. Prostatitis, with evidence found in PMIDs: 34428489, 29194865, 33814544, 29605575, 33243958, 30145778, 37857076, 31587256, 31689885, and 31831298. [Hallucination: PMIDs: 33814544, 29605575, 33243958, 30145778, 37857076 and 31689885 are related to prostate diseases but not prostatitis.]<br><br>5. Lower Urinary Tract Symptoms, with evidence found in PMIDs: 34532268, 37681541, 34817584, 29665997, 29379733, 32202088, 30446471, 30415110, 30552307, and 31179381.<br><br>Please note that these are the PMIDs of the articles that contain the evidence. It's recommended to read the full articles for comprehensive information. |